# Boundary-semantic collaborative guidance network with dual-stream feedback mechanism for salient object detection in optical remote sensing imagery

Dejun Feng, Hongyu Chen, Suning Liu, Ziyang Liao, Xingyu Shen, Yakun Xie, Jun Zhu

*Abstract*—With the increasing application of deep learning in various domains, salient object detection in optical remote sensing images (ORSI-SOD) has attracted significant attention. However, most existing ORSI-SOD methods predominantly rely on local information from low-level features to infer salient boundary cues and supervise them using boundary ground truth, but fail to sufficiently optimize and protect the local information, and almost all approaches ignore the potential advantages offered by the last layer of the decoder to maintain the integrity of saliency maps. To address these issues, we propose a novel method named boundary-semantic collaborative guidance network (BSCGNet) with dual-stream feedback mechanism. First, we propose a boundary protection calibration (BPC) module, which effectively reduces the loss of edge position information during forward propagation and suppresses noise in low-level features without relying on boundary ground truth. Second, based on the BPC module, a dual feature feedback complementary (DFFC) module is proposed, which aggregates boundary-semantic dual features and provides effective feedback to coordinate features across different layers, thereby enhancing cross-scale knowledge communication. Finally, to obtain more complete saliency maps, we consider the uniqueness of the last layer of the decoder for the first time and propose the adaptive feedback refinement (AFR) module, which further refines feature representation and eliminates differences between features through a unique feedback mechanism. Extensive experiments on three benchmark datasets demonstrate that BSCGNet exhibits distinct advantages in challenging scenarios and outperforms the 17 state-of-the-art (SOTA) approaches proposed in recent years. Codes and results have been released on GitHub: https://github.com/YUHsss/BSCGNet.

*Index Terms*—Salient object detection, Optical remote sensing image, Boundary-Semantic, Collaborative guidance, Feature feedback.

## I. Introduction

Salient object detection (SOD) is a technique that aims to emulate the human visual system by identifying the most

This work was supported in part by the National Key Research and Development Program of China (Grant No. 2022YFC3005703), in part by the Postdoctoral Innovation Talents Support Program (Grant No. BX20230299), in part by the National Natural Science Foundation of China (Grant Nos. 42271424, 42171397 and 42301473). *(Corresponding author: Hongyu Chen)*

Dejun Feng is with the Faculty of Geosciences and Environmental Engineering, Southwest Jiaotong University, Chengdu 610097, China, also with the State-Province Joint Engineering Laboratory of Spatial Information Technology of High-Speed Rail Safety, Southwest Jiaotong University, Chengdu 610031, Sichuan, China (e-mail: djfeng@swjtu.edu.cn)

Hongyu Chen, Suning Liu, Ziyang Liao, Xingyu Shen, Yakun Xie and Jun Zhu are with the Faculty of Geosciences and Environmental Engineering, Southwest Jiaotong University, Chengdu 610097, China (e-mail: chy0519@my.swjtu.edu.cn; sakura_ningning@163.com; LZY2021@my.swjtu.edu.cn; sxyu@my.swjtu.edu.cn; yakunxie@163.com; zhujun@swjtu.edu.cn)

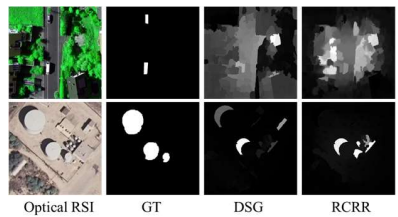

Fig. 1. Illustration of the performance of traditional methods.

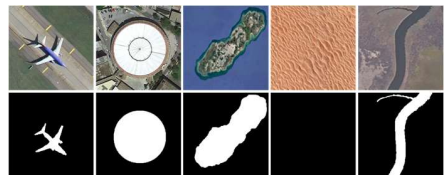

Fig. 2. Examples of optical RSIs: (top row) optical RSIs and (bottom row) ground truth (GT).

visually salient regions within a given scene. It has been widely applied in various domains, including natural images [1], videos [2], RGB-D images [3], and has found success in target tracking [4], [5], quality assessment [6], [7], and image segmentation [8], [9]. Moreover, with the improvement of the spatial resolution of optical remote sensing images (ORSIs), SOD holds practical value in remote sensing applications, such as burn area extraction [10], real-time detection [11], and more. However, unlike SOD in natural scene images (NSIs), ORSIs pose unique challenges due to its larger coverage area, complex backgrounds, numerous ground objects, and diverse types and scales of target objects. Therefore, achieving accurate salient object detection in RSIs is a challenging task, essential for enhancing image analysis and understanding.

Over the past few decades, a multitude of methods have been proposed for accurate SOD to identify visually prominent regions within an image. These methods can be classified into two distinct categories: traditional methods and deep learning (DL) methods. Traditional methods can be subdivided into two categories: handcrafted feature-based methods and machine learning-based methods. Handcrafted feature-based methods mostly rely on classical principles to create manual features, such as sparse representation [12], [13], feature fusion [14], and color information content [15]. While these methods have a strong theoretical foundation, they often struggle to accurately detect target objects in scenes with rich and complex features. The presence of noise surrounding the target object can lead to inaccurate detection results. With the accumulation of data, machine learning methods have been proposed by many researchers, which aim to obtain detection results by training classifiers such as random walks [16], [17], support vector machine [18], [19], conditional random field [20], [21], etc. Despite the superior efficiency of machine learning methods over handcrafted feature-based methods, they may encounter limitations in effectively learning deep-level features and adapting to the complexity of ORSI. Fig. 1 demonstrates that both traditional methods struggled to precisely focus on interest regions when confronted with ORSIs characterized by complex backgrounds.

In recent years, the rapid development of DL methods, particularly those based on convolutional neural networks (CNNs), has led to significant advancements in addressing the challenges posed by complex target structures and scenes in images. CNNs excel in automatically extracting and learning image features and enabling effective feature interactions [79-84]. As a result, numerous CNN-based models have been introduced into NSI-SOD, leveraging these capabilities [22-25]. CNNs play a pivotal role in improving the accuracy and driving the advancement of SOD. However, NSI and ORSI exhibit substantial disparities. NSIs are typically captured using handheld cameras by humans, while ORSIs are acquired from high-angle perspectives using satellites or aircraft. Consequently, compared with NSIs, ORSIs present larger coverage areas and significant variations in target scale, complexity, orientation, and brightness. Furthermore, ORSIs may even lack salient targets. Some representative scenarios are shown in Fig. 2. Directly migrating NSI-SOD methods to ORSI can lead to inaccurate detection results. Recognizing this challenge and with the emergence of an expanding collection of ORSI-SOD public datasets, numerous researchers have devised CNN models specifically tailored for ORSI-SOD. LVNet [9], one of the early works in ORSI-SOD, uses nested connections to link detail and semantic features to suppress the interference of complex backgrounds. Subsequently, edge guidance [26]-[31], attention [4], [32]-[34], and feature fusion [26]-[31] have also been applied to ORSI-SOD to improve detection accuracy. Subsequently, the emergence of the transformer model, known for its strong ability to capture global information [36], led to the development of transformer-based ORSI-SOD models [29], [30], [37]. However, the high computational cost and memory requirements associated with transformer models impose limitations on their scalability for processing large volumes of information in remote sensing scenario. Consequently, the utilization of lightweight models in ORSI-SOD has been proposed as a solution to mitigate this issue [26], [38]. However, it is important to note that lightweight models, despite their advantages, may encounter challenges in capturing the intricate features and relationships within the data due to their limited parameterization. This limitation could result in reduced performance in challenging scenarios and constraints on their generalization capabilities.

While CNN-based ORSI-SOD methods have achieved significant improvements in detection accuracy, a more in-depth analysis has unveiled certain limitations in these approaches.

1) *Regarding the handling of boundaries:* Several existing methods [37], [39], [40] do not pay sufficient attention to the influence of boundary contour information on detection results, resulting in blurred boundaries in the output. To capture detailed boundary information, both existing ORSI-SOD and NSI-SOD methods [41]-[43] commonly employ the output of encoding layers as direct boundary constraint information, often without adequate protection and optimization. Alternatively, some methods incorporate boundary information into the loss function. Moreover, while high-level features contain rich semantic information, they often lack precise positional information. Hence, how to complement the location information and semantic information among layers, to make each layer contain multi-level features remains a challenge that existing methods need to address.

2) *Regarding the exploration of the last layer of the decoder:* Almost all existing methods [26], [28], [35], [40], [44], [45] fail to adequately acknowledge the significance of the last decoding layer and capitalize on its full potential. These methods typically directly aggregate information from different layers or even generate a single output for generating a saliency map. In particular, the last layer is typically in closer proximity to the GT. By utilizing it to optimize the missing features across different layers, a more comprehensive contextual information and accurate feature representation can be achieved.

After the above detailed analysis, we propose a novel method dedicated to ORSI-SOD called boundary-semantic collaborative guidance network (BSCGNet) with dual-stream feedback mechanism. To effectively preserve boundary information within low-level features and facilitate the aggregation of multi-scale information to mitigate the semantic gap between features of varying scales, BSCGNet addresses existing challenges through the adoption of stepwise calibration and dual-feedback strategies. These approaches significantly depart from previous ORSI-SOD methods. The principal contributions are summarized as follows.

1) We propose a novel BSCGNet dedicated to ORSI-SOD, which explores detail edge features and deep semantic features in a unique design. To the best of our knowledge, this manner is different from existing works that mostly focus on the extraction and fusion of boundary and semantic features from salient objects in RSIs. Extensive comparative experiments and analyses have been conducted, demonstrating that BSCGNet surpasses 17 other state-of-the-art (SOTA) methods on three publicly available datasets. Meanwhile, the effectiveness and rationality of the proposed module are further validated through detailed visualization of feature maps.

2) We designed a boundary protection calibration (BPC) module for the encoder, which comprises two sub-modules. The first sub-module is the adjacent feature attention (AFA) sub-module, which integrates adjacent features to enhance the representation of boundary location features. The second sub-module is the feature calibration (FC) sub-module, which gradually calibrates low-level features to emphasize edge clues. To mitigate the significant loss of edge information during forward propagation, we adopt a progressive optimization approach.

3) Two unique feedback mechanisms are introduced to the encoder and decoder. In the encoder, the dual feature feedback

complementary (DFFC) module feeds back the captured boundary-semantic dual features to coordinate and complement features at all levels. In the decoder, we propose the adaptive feedback refinement (AFR) module, which aims to reduce inter-feature discrepancies and provide valuable information for obtaining a more refined saliency map. To the best of our knowledge, this is the first time that the uniqueness of the last layer of the decoder is considered in the design of the ORSI-SOD method.

4) To enhance further research in SOD, we offer a shared code repository containing 17 SOD methods, including 8 ORSI-SOD methods, 7 NSI-SOD methods, 2 traditional SOD methods and the method proposed in this paper.

## II. RELATED WORK

In this section, we briefly introduce the current status of SOD and the existing methods for exploring boundary-semantics.

### A. Current Status of SOD

SOD was initially applied in NSIs. Itti et al. [46] proposed the center-surround difference theory, which was a groundbreaking work, and numerous theory-based methods were subsequently proposed. For example, Ren et al. [47] integrated region contrast, depth, and orientation priors to achieve SOD. Wang et al. [48] proposed a supervised multiple instances learning framework for SOD and utilized low-, mid-, and high-level features for testing. Wei et al. [49] used background priors in NSI to provide more clues for SOD. With the maturity of technology, numerous traditional machine learning methods have also been applied in this field. For instance, Liu et al. [50] used conditional random fields to combine numerous features for SOD, and Jiang et al. [51] used the random forest method to map feature vectors to saliency scores. In the work of Zhou et al. [19], an SVM-based method was proposed to generate complementary saliency maps. Similarly, in the work of Tong et al. [18], multiple SVM results were integrated to obtain the final saliency result. Although these traditional methods have strong theoretical foundations, most of them are based on handcrafted features and often fail in complex and cluttered scenes.

In recent years, in view of the advantages of DL methods in the field of computer vision, more and more researchers have applied it to SOD, significantly promoting the development of the SOD. Li and Yu [52] used three convolutional branches to extract multi-scale depth features on images with different resolutions, optimizing the saliency results and spatial consistency. This was the pioneering work of applying DL methods to SOD. Subsequently, Deng et al. [53] proposed R3Net, which uses residuals to obtain complementary saliency information and refined the saliency map with residuals to obtain more accurate salient regions. Hu et al. [54] proposed a fully convolutional network (FCN) with cyclic aggregated deep features for salient object detection, using multi-level features in a recursive manner to gradually refine the depth features of each layer. Wu et al. [55] proposed a decoder that discarded larger-resolution features from shallow layers and directly used the generated saliency map to optimize the network, effectively suppressing interference. Chen et al. [56] used a discriminative cross-modal transfer learning network for RGB-D SOD. In addition, edge details in the saliency map are important information, and now many models consider using edge information as auxiliary information to obtain fine contour details. In the work of Qin et al. [57], BASNet, a prediction-refinement architecture, and a mixed loss focusing on edges were proposed to improve attention to edges in the saliency map. Sun et al. [58] proposed BGNet to explore object-related edge information. Although these DL-based methods can achieve good results in SOD, there are certain differences between NSI and ORSI, and if forced migration is carried out, it will often backfire.

As a result of the importance of ORSI and SOD, the research of ORSI-SOD is gradually increasing. The traditional methods of ORSI-SOD are similar to unsupervised NSI-SOD models. For example, In the work of Zhao et al. [12], sparse representation was used to obtain global and background information for generating saliency maps. Ma et al. [59] yielded feature maps using structure tensors and background contrast, which were then fused to produce super pixel saliency map. Liu et al. [60] constructed a colored Markov chain and combined the latent saliency map with circular feature maps to generate the best saliency map. However, it is worth noting that due to the rapid development of DL, the traditional ORSI-SOD method is becoming less and less.

Driven by the increasing adoption of DL-based methods, numerous researchers have conducted extensive investigations into DL-based ORSI-SOD approaches. In the early stages, due to the limited availability of ORSI datasets, many researchers employed weakly supervised learning techniques to tackle RSI-SOD challenges [61], [62]. With the emergence of RSI datasets, more and more researchers have begun to explore SOD models specifically for remote sensing scenarios. For example, Zhou et al. [27] utilize multi-scale feature ensembles to highlight boundary cues and utilize boundary GT for supervised learning. Li et al. [26] model encoder low-level features and employ attention mechanisms for details enhanced. Wang et al. [65] build a semantic guidance scheme to guide the model to obtain fine-grained semantic feature information; Zhen et al. [30] perceive global-local semantics by modeling contextual information.

### B. Existing Methods for Exploring Boundary-Semantics

To obtain saliency maps with fine contours, most existing DL-based ORSI-SOD methods rely on boundary information in low-level features and build attention mechanisms to explore boundary cues [26]-[31]. In addition, to capture sufficient semantic information of salient objects, many methods adopt feature fusion strategies [26], [28], [30], [32], [64], [65].

In general, the above approaches for exploring boundary information and capturing fine-grained semantics do not deal well with objects in challenging scenes. However, in our proposed method BSCGNet, we employ a progressive calibration strategy within the encoder to refine the boundaries. This strategy can effectively protect the loss of boundary information during forward transmission and preserve boundary-semantic dual features, all without relying on any boundary labels for supervised learning. By doing so, we ensure that the boundary information is preserved and not significantly lost during forward propagation, thereby providing a more accurate foundation for generating the saliency map. When capturing fine-grained semantics, we focus on the uniqueness of the last layer in the decoder to enhance multi-scale representation. This enables us to capture finer details in the semantics and obtain a more comprehensive and detailed saliency map. It is worth noting that these handling of boundaries and semantics are completely different from the above methods, like SeaNet [26], BANet [45], GLGCNet [30], JRBM [42], EMFINet [27] and HFANet [35]. In conclusion, our BSCGNet is a novel approach for RSI-SOD with superior performance in exploring boundary-semantic information in challenging scenarios.

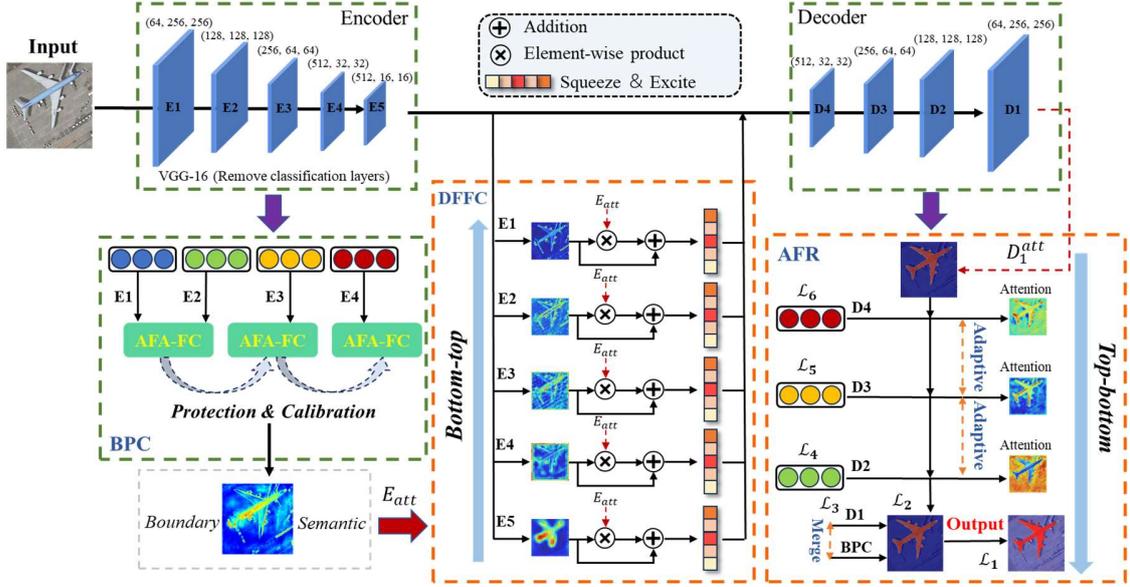

**Fig. 3.** The overall architecture of the proposed BSCGNet.

## III. METHODOLOGY

### A. Overview

The overall framework of BSCGNet, as illustrated in Fig. 3, consists of five key components: Encoder, BPC module, DFFC module, decoder, and AFR module. For feature extraction, we adopt VGG16 [66] as the fundamental backbone, omitting its final four layers primarily employed for classification purposes. If the input image is denoted as $E \in \mathbb{R}^{3 \times H \times W}$, it is transformed to $E \in \mathbb{R}^{512 \times \frac{H}{16} \times \frac{W}{16}}$ after passing through the backbone. Then, the BPC module is utilized to safeguard boundary information during the backward propagation process and learn the offset of information between adjacent layers. This module effectively calibrates the position information of boundaries. Subsequently, the DFFC module is employed to supplement features from each layer of the encoder. This ensures that the features contain both rich semantics and accurate boundary information. Lastly, to refine the saliency map further, the AFR module is employed to optimize and rectify the multi-level features from the decoder. In the subsequent sections, we will introduce each key component in detail and fully verify their rationality and effectiveness in experiments.

### B. Boundary Protection Calibration Module

To minimize the loss of boundary position information during forward propagation and mitigate the interference of noise in low-level features, we propose the BPC module. The BPC module aims to progressively protect and optimize all levels of the encoding layer, mitigating the loss of boundary information from shallow to deep layers. The overall framework of the BPC module is depicted in Fig. 4. We will elaborate on the module in three steps.

*1) AFA sub-module:* Taking the first two layers of the encoder as an example, first, we upsample the first two layers of adjacent feature map of the encoder (i.e., $E_i \{i = 1, i = 2\}$) by a bilinear interpolation method to obtain $E_i^{up} \in \mathbb{R}^{3 \times H \times W} \{i = 1, i = 2\}$. Second, $E_i^{up} \{i = 1\}$ is optimized through a residual module consisting of two convolutional blocks, and the captured features can be denoted as $f_i^{res} \{i = 1\}$. Finally, the adjacent feature layer information will be integrated by concatenating $f_i^{res} \{i = 1\}$ and $E_i^{up} \{i = 2\}$, and the perception and attention of low-level feature information are strengthened by using the SELayer [67] to strengthen the network's focus on more important regions. The entire process can be described as.

$$f_a = SE(Cat(Res(Up(E_1)), Up(E_2))) \quad (1)$$

where $f_a$ represents the output feature map $f_a \in \mathbb{R}^{128 \times H \times W}$; $SE(\cdot)$ denotes the SELayer; $Cat(\cdot)$ denotes the concatenate operation; $Res(\cdot)$ denotes the residual module; $Up(\cdot)$ denotes the upsampling operation; $E_1$, $E_2$ represent the first two layers of VGG16.

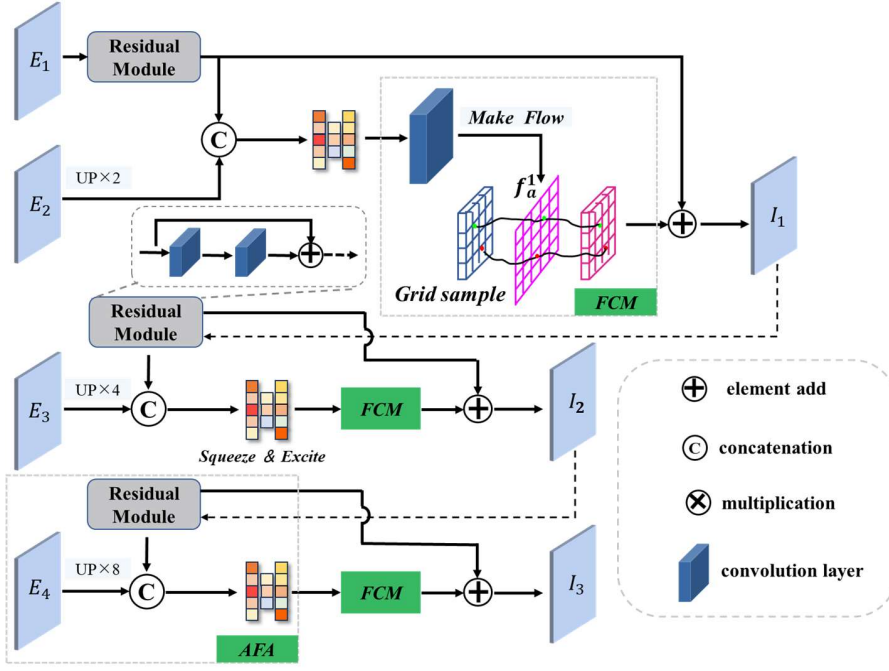

**Fig. 4.** Architecture of the proposed BPC module.

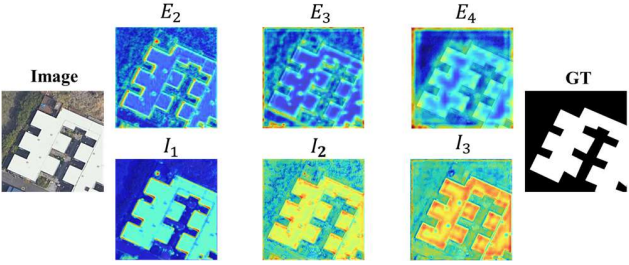

**Fig. 5.** Visualization of encoder step-by-step calibration.

*2) FC sub-module:* To better protect the encoder boundary location information and calibrate the target information step by step to reduce the interference of noise, we try to obtain the semantic offsets between adjacent feature layers. First, based on the obtained adjacent feature information $f_a$, the number of channels is further reduced to 2 by using a 3×3 convolution to obtain the semantic offset $f_a^1 \in \mathbb{R}^{2 \times H \times W}$. It is worth noting that $f_a^1$ is obtained by calculating the flow field between the adjacent feature layers (i.e., $f_1^{res}$ and $E_2^{up}$). The process can be represented as follows.

$$f_a^1 = Conv_{3\times 3}(f_a) \qquad (2)$$

where $Conv_{3\times 3}(\cdot)$ represents the 3×3 convolution operation. Then, based on the obtained semantic offset $f_a^1$, we can iterate through each pixel in $E_2^{up}$ and remap it to a grid of the same size as the Flow Field. Afterwards, $f_1^{res}$ will be fused with more accurate features generated by remapping. Through this process, the network can perform more accurate feature propagation and gradually correct the boundary position information captured by shallow layers. The calibration process can be described as.

$$I_1 = GS(E_2^{up}, f_a^1) \oplus f_1^{res} \qquad (3)$$

where $GS(\cdot)$ represents grid sampling, it guides the model's attention towards regions of interest by employing pixel-level mappings between two layers, enhancing the original segmentation features while preserving boundary cues to the greatest extent possible. This alleviates information loss during the forward propagation process, facilitating improved transmission of boundary information across layers, ultimately leading to enhanced detection accuracy; $E_2^{up}$ represents the upsampled result of the second feature layer in the encoder; $f_a^1$ is the semantic offset; $f_1^{res}$ represents the result of the first feature layer in the encoder refined by residual blocks, and $\oplus$ denotes the element-wise addition.

*3) Level-by-level optimization:* Building upon (1) and (2), $I_1$ is fed into the same residual block as that in (1), and $E_i$ $\{i = 3\}$ is upsampled to obtain $E_i^{up}$ $\{i = 3\}$. The information from both is integrated, and the flow field between $f_{I_1}^{res}$ and $E_3^{up}$ is computed using the same approach to obtain the semantic offset. This offset is then remapped to correct the features of $E_3$ and $E_4$, similar to the process between $f_{I_1}^{res}$ and $E_3^{up}$. With this level-by-level optimization, more precise boundary information can be obtained. The comparison between the outputs before and after applying the BPC module can be observed in Fig. 5. During the forward propagation, the BPC module effectively preserves boundary information while enhancing the semantic representation of salient object interiors. It is noteworthy that only the first four layers of the encoder are used for optimization since the boundary position information is more apparent in these layers compared with the last layer. The process of level-by-level optimization can be expressed as.

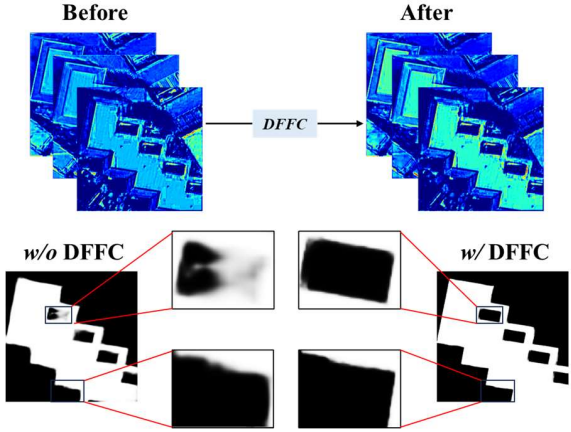

**Fig. 6.** The DFFC module's treatment of internal features and boundary details.

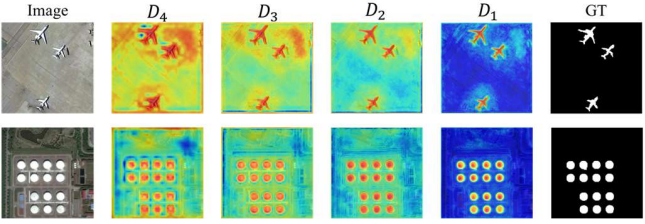

**Fig. 7.** Visualization of the layers of the decoder.

$$\begin{cases} I_1 = GS(E_2^{up}, f_a^1) \oplus f_1^{res} \\ I_2 = GS(E_3^{up}, f_a^2) \oplus I_1 \\ I_3 = GS(E_4^{up}, f_a^3) \oplus I_2 \end{cases} \quad (4)$$

where $I_i\{i=1.2.3\}$ denotes the results after correction; $E_i^{up}\{i=2,3,4\}$ denotes the corresponding upsampling operation of feature layers; $f_a^i\{i=1,2,3\}$ represents the semantic offset between adjacent layers after (1); $f_1^{res}$ denotes the refined result of the first feature layer in the encoder by residual blocks. The saliency map outputted by the BPC module is finally denoted as $\mathcal{M}_1 \in \mathbb{R}^{64 \times H \times W}$.

### C. Dual Feature Feedback Complementary Module

The last layer of the encoder is usually the global perception of semantic features. Therefore, based on the accurate boundary information captured by the BPC module, boundary-semantic dual features can be used to supplement the feature information of each layer of the encoder. To achieve this, the DFFC module is proposed, the details of which have been presented in Fig. 3.

First, the boundary information $I_3$ corrected in section III(B) 3) is fused with the feature information of the last encoder layer $E_5$. Subsequently, the boundary-semantic dual-feature attention map is obtained through the sigmoid function. This process can be expressed as.

$$\mathcal{P} = Cat(I_3, Up(Conv_{1\times1}(E_5))) \quad (5)$$
$$\mathcal{P}' = \sigma(Conv_{1\times1}(\mathcal{P})) \quad (6)$$

where $Conv_{1\times1}(\cdot)$ denotes a 1×1 convolution, and $\sigma(\cdot)$ represents the sigmoid function.

Finally, to enable efficient feedback transfer of the attention map, the attention map $\mathcal{P}'$ (i.e., $E_{att}$ in Fig. 3) is directly downsampled to the same size as each feature maps in the encoder before multiplying and adding. This operation is used to effectively complement the features of each layer. Besides, the SELayer, which enriches the dual-feature information (i.e., semantic and boundary information) in layers of encoder, is used after each feedback to strengthen attention to important information. The entire process is expressed as follows.

$$G_i = SE((DS(\mathcal{P}') \oplus 1) \otimes E_i) \quad (7)$$

where $E_i$ represents the feature maps of the encoder layers, with $i = 5,...,1$; $DS(\cdot)$ denotes down-sampling; $\otimes$ denotes element-wise multiplication. Subsequently, we decode $G_i(i = 5,...,1)$ using transposed convolution and neighboring layer fusion.

To demonstrate more clearly that the DFFC module can effectively enhance the complementarity of boundary-semantic dual features, we conducted a preliminary visualization analysis. Taking the first layer of the encoder as an example (i.e., $E_1$), Fig. 6 illustrates the changes in representative feature maps along the channel dimensions for $E_1$ before and after passing through the DFFC module, along with the handling of boundary details with (*w/*) and without (*w/o*) DFFC module. It is evident from these observations that the DFFC module not only enhances the internal features of prominent objects but also effectively refines the boundary information.

### D. Adaptive Feedback Refinement Module

Most of the current SOD methods ignore the importance of the last layer in the decoder. Theoretically, the output of the last layer of the decoder is usually closest to the GT, which can be clearly found from Fig. 7. Although the BPC module and DFFC module can make the feature map information of different layers relatively diverse, these two modules are not capable of completely eliminating the redundant information and noise produced by the encoder. Therefore, to obtain more accurate saliency maps and effectively alleviate these problems, we introduce a novel AFR module, and the details are shown in Fig. 8.

*1) Feedback details:* To obtain more complete features at each level of the decoder and output more accurate saliency map, we first directly downsample the last decode layer to the same size as the other three layers to obtain three feedback features: $\mathcal{FB}_4 \in \mathbb{R}^{64 \times \frac{H}{8} \times \frac{W}{8}}$, $\mathcal{FB}_3 \in \mathbb{R}^{64 \times \frac{H}{4} \times \frac{W}{4}}$, and $\mathcal{FB}_2 \in \mathbb{R}^{64 \times \frac{H}{2} \times \frac{W}{2}}$, and refine the features of each layer by feeding forward in turn. Simply put, $\mathcal{FB}_i\{i=4,3,2\}$ are integrated with the corresponding feature maps of the same size, which can be described as.

$$\mathcal{F}_i = \mathcal{FB}_i + Conv_{1\times1}(\mathcal{D}_i)\ \{i=4,3,2\} \quad (8)$$

where $\mathcal{F}_i$ represents the preliminary feature maps after feedback, $\mathcal{FB}_i$ represents the downsampled feedback feature maps, $\mathcal{D}_i$ represents the other three layers of the decoder.

*2) Inter-layer information refinement (IIR) sub-module:*

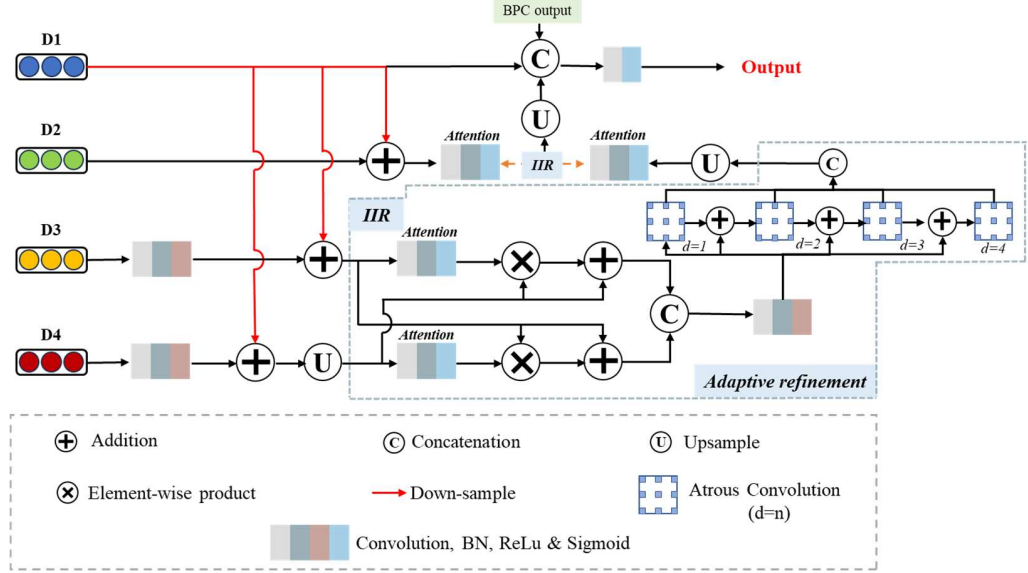

**Fig. 8.** Architecture of the proposed AFR module.

Considering that there are some differences in the information contained between the layers and to capture the rich contextual information between different scales and suppress the inter-feature differences. First, we generate attention maps $\mathcal{A}_4$ and $\mathcal{A}_3$ by applying the sigmoid function to $\mathcal{F}_4$ and $\mathcal{F}_3$, respectively. Mathematically, this process can be expressed as follows.

$$\mathcal{A}_i = \sigma(Conv_{3\times3}(Conv_{3\times3}(\mathcal{F}_i))) \ \{i = 1,2\} \quad (9)$$

where $\mathcal{F}_i$ denotes the preliminary feature maps obtained after feedback.

Next, the features of each layer are adaptively complemented and refined by cross-multiplying and adding to minimize the feature differences between the various layers prior to merging. This process can be described as follows.

$$\mathcal{C}_1 = Conv_{3\times3}((\mathcal{A}_4 \oplus 1) \otimes Conv_{3\times3}(\mathcal{F}_3)) \quad (10)$$
$$\mathcal{C}_2 = Conv_{3\times3}((\mathcal{A}_3 \oplus 1) \otimes Conv_{3\times3}(\mathcal{F}_4)) \quad (11)$$
$$\mathcal{S} = Conv_{1\times1}(Cat(\mathcal{C}_1, \mathcal{C}_2)) \quad (12)$$

where $\mathcal{C}_i\{i = 1,2\}$ represents the cross-refined feature maps, $\mathcal{A}_i\{i = 4,3\}$ represents attention maps, $\mathcal{F}_i\{i = 4,3\}$ denotes preliminary feature maps after feedback, and S represents the merged feature map.

Finally, as depicted in Fig. 8, a dense connectivity approach is employed to capture contextual information at various scales. Four dilated convolutions are parallelly connected with increasing dilation rates (1, 2, 3, and 4). The output of each dilated convolution is element-wise added to the original feature mapping before being passed to the subsequent dilated convolution. This stepwise expansion of the receptive field gradually extended the network's "visual scope" to encompass a broader range. Notably, for $\mathcal{D}_2$, we cross-learn it with the results of the first two feedback refinements in a process consistent with that described above. The decoder layers are gradually refined with these three rounds of feedback optimization, resulting in more accurate saliency maps.

*3) Information fusion and deep supervision strategy:* Further information fusion is performed to improve the completeness of the final saliency map. Specifically, we integrate the last layer of the decoder, the boundary information $I_3$ corrected in section III(B) 3), and the refined results obtained by semantic feedback. Then, we obtain the final output through a 1×1 convolution. Moreover, a deep supervision strategy [68] is adopted to ease the training of the shallow layers and mitigate gradient explosion or disappearance. More precisely, we use a 1×1 convolution to generate an output at the end of each of the four decoder stages and after section III(D) 2) process.

*E. Hybrid Loss Functions*

To enhance the network's representation of multi-source features, a hybrid loss function is employed in this study. It is defined as follows.

$$\ell_{joint} = \ell_{BCE} + \ell_{IoU} \quad (13)$$

where $\ell_{BCE}$ and $\ell_{IoU}$ denote the binary cross-entropy (BCE) loss and the IoU loss, respectively.

$\ell_{BCE}$ is defined as.

$$\ell_{BCE} = -\frac{1}{N}\sum_{i=1}^{N}[y_i \log p_i + (1-y_i)\log(1-p_i)] \quad (14)$$

where $p_i$ is the prediction mapping and $y_i$ is the true label, and $N$ represents the number of pixels in the input image.

$\ell_{IoU}$ is defined as.

$$\ell_{IoU} = 1 - \frac{\sum_{i=1}^{N} p_i y_i}{\sum_{i=1}^{N}(p_i + y_i - p_i y_i)} \quad (15)$$

The meaning of each of these symbols is the same as $\ell_{BCE}$.

## IV. EXPERIMENTS

*A. Datasets and Implementation Details*

*1) Datasets:* The proposed method is trained and tested on three publicly available ORSI-SOD datasets.

**ORSSD** [9]: The dataset consists of 800 ORSIs with corresponding pixel-level annotations, including 600 training

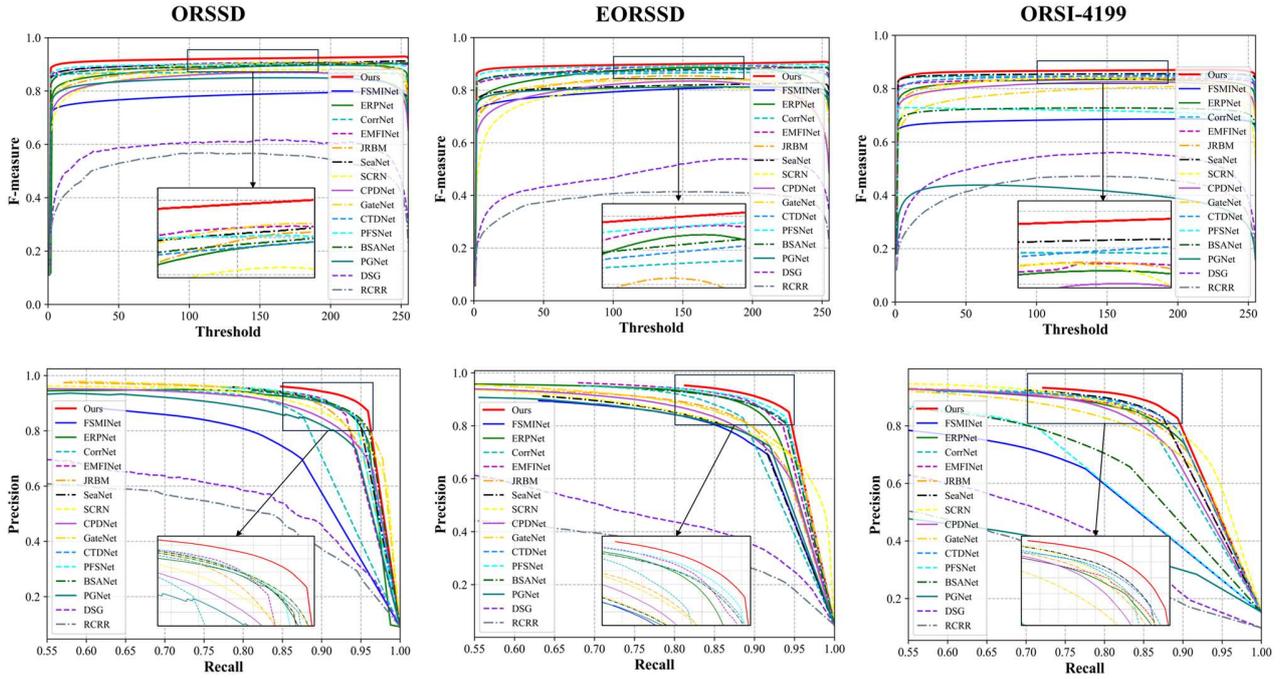

**Fig. 9.** Quantitative comparisons of different saliency models. The first and second rows are F-measure and P-R curves, respectively. *Please zoom-in for the best view.*

TABLE I

QUANTITATIVE COMPARISON RESULTS ON THE ORSSD, EORSSD AND ORSI-4199 DATASETS. HERE, "↑" (↓) MEANS THAT THE LARGER (SMALLER) THE BETTER. THE BEST THREE RESULTS IN EACH ROW ARE MARKED IN **RED**, **BLUE**, AND **GREEN**, RESPECTIVELY

| Methods | Publication | FLOPs (G) | Params. (M) | Model size (M) | FPS | ORSSD (200 images) | | | EORSSD (600 images) | | | ORSI-4199 (2199 images) | | |
|---|---|---|---|---|---|---|---|---|---|---|---|---|---|---|
| | | | | | | $\mathcal{M} \downarrow$ | $S_\alpha \uparrow$ | $F_\beta \uparrow$ | $\mathcal{M} \downarrow$ | $S_\alpha \uparrow$ | $F_\beta \uparrow$ | $\mathcal{M} \downarrow$ | $S_\alpha \uparrow$ | $F_\beta \uparrow$ |
| | | | | | | Traditional methods | | | | | | | | |
| DSG | TIP 2017 | - | - | | - | 0.1024 | 0.7186 | 0.6090 | 0.1250 | 0.6423 | 0.5233 | 0.1294 | 0.6971 | 0.6270 |
| RCNN | TIP 2018 | - | - | | - | 0.1277 | 0.6849 | 0.5591 | 0.1647 | 0.6011 | 0.4018 | 0.1637 | 0.6491 | 0.5480 |
| | | | | | | DL-based NSI-SOD methods | | | | | | | | |
| SCRN | ICCV 2019 | 7.97 | 25.23 | 94 | 50 | 0.0166 | 0.8951 | 0.8623 | 0.0174 | 0.8490 | 0.7898 | **0.0294** | 0.8689 | 0.8484 |
| CPDNet | CVPR 2019 | 31.44 | 29.23 | 112 | 102 | 0.0187 | 0.8954 | 0.8524 | 0.0111 | 0.8874 | 0.8093 | **0.0281** | 0.8634 | 0.8343 |
| GateNet | ECCV 2020 | 96.21 | 100.02 | 382 | 110 | 0.0131 | 0.9074 | 0.8913 | 0.0130 | 0.8719 | 0.8157 | 0.0378 | 0.8425 | 0.8169 |
| CTDNet | ACM 2021 | 3.24 | 11.82 | 45 | 212 | 0.0138 | 0.9197 | 0.8853 | 0.0088 | 0.9190 | 0.8610 | 0.0312 | **0.8727** | **0.8617** |
| PFSNet | AAAI 2021 | 24.02 | 31.18 | 119 | 68 | 0.0149 | 0.9187 | 0.8832 | **0.0079** | 0.9233 | 0.8705 | 0.0791 | 0.7596 | 0.7372 |
| BSANet | AAAI 2022 | 13.19 | 28.79 | 125 | 38 | 0.0119 | 0.9170 | 0.8873 | **0.0080** | 0.9182 | 0.8610 | 0.0471 | 0.7965 | 0.7342 |
| PGNet | CVPR 2022 | 18.37 | 72.62 | 279 | 43 | 0.0225 | 0.8894 | 0.8338 | 0.0153 | 0.8757 | 0.7887 | 0.2177 | 0.5683 | 0.4371 |
| | | | | | | DL-based ORSI-SOD methods | | | | | | | | |
| LVNet | TGRS 2019 | - | - | - | - | 0.0207 | 0.8815 | 0.8263 | 0.0146 | 0.8630 | 0.7794 | - | - | - |
| FSMINet | GRSL2022 | 5.24 | 3.56 | 14 | 54 | 0.0101 | 0.9361 | 0.9041 | **0.0079** | 0.9255 | 0.8678 | 0.0789 | 0.7557 | 0.6946 |
| ERPNet | TCYB 2022 | 113.89 | 56.48 | 216 | 58 | 0.0135 | 0.9254 | 0.8974 | 0.0082 | 0.9210 | 0.8632 | 0.0305 | 0.8633 | 0.8436 |
| CorrNet | TGRS 2022 | 21.08 | 4.07 | 16 | 74 | **0.0098** | **0.9380** | **0.9129** | 0.0083 | **0.9289** | **0.8778** | 0.0366 | 0.8623 | 0.8560 |
| EMFINet | TGRS 2022 | 176.67 | 95.09 | 363 | 39 | 0.0109 | **0.9366** | 0.9002 | 0.0084 | **0.9290** | 0.8720 | 0.0376 | 0.8664 | 0.8486 |
| JRBM | TGRS 2022 | 50.71 | 43.54 | 167 | 55 | 0.0163 | 0.9204 | 0.8842 | 0.0099 | 0.9197 | 0.8656 | 0.0374 | 0.8593 | 0.8493 |
| SRAL | TGRS 2023 | - | - | - | - | **0.0105** | 0.9305 | **0.9167** | **0.0067** | 0.9234 | **0.8964** | 0.0321 | **0.8735** | 0.8576 |
| SeaNet | TGRS 2023 | 1.35 | 2.74 | 11 | 99 | **0.0105** | 0.9260 | 0.8942 | 0.0083 | 0.9208 | 0.8649 | **0.0308** | 0.8722 | **0.8653** |
| Ours | Submission | 86.35 | 26.99 | 103 | 60 | **0.0093** | **0.9411** | **0.9101** | **0.0079** | **0.9296** | **0.8808** | **0.0241** | **0.8892** | **0.8797** |

images and 200 testing images.

**EORSSD** [63]: The dataset expands the original ORSSD dataset to 2000 ORSIs with corresponding pixel-level GT. It contains 1400 training images with corresponding labels and 600 testing images with corresponding labels.

**ORSI-4199** [42]: The ORSI-4199 dataset is the biggest and most challenging ORSI-SOD dataset. It contains 2000 images for training and 2199 images for testing, in which nine attributes are annotated to facilitate analyzing the strengths and weaknesses of SOD models from different perspectives.

*2) Implementation Details:* All experiments are conducted on a server equipped with an NVIDIA GeForce RTX 3090 24G. We used the Adam optimizer with a learning rate of 1e-4 and a batch size of 8. The models are trained for 50 epochs on the ORSSD dataset, 45 epochs on the EORSSD dataset, and 35 epochs on the ORSI-4199 dataset, with the learning rate multiplied by 0.1 after every 30th epoch. In addition, each image is resized to 256 × 256 before inputting due to memory limitations. To prevent over-fitting during the training process, we adopt a series of data augmentation strategies, such as random rotation, horizontal, vertical flipping, and Gaussian blurring.

### B. Evaluation metrics

We used three commonly used SOD metrics to evaluate the performance of all the methods, including Mean Absolute Error Value ($\mathcal{M}$) [76], S-measure Value ($S_\alpha$) [77], and max $F_\beta$ [78]. Mathematical formulations for each evaluation metric are presented below.

$$\mathcal{M} = \frac{1}{l \times w} \sum_{i=1}^{l} \sum_{j=1}^{w} |S(i,j) - G(i,j)| \quad (16)$$

It represents the mean pixelwise error between the predicted salient map ($\mathcal{M}$) with the GT ($G$), where $l$ and $w$ denote the length and width of $\mathcal{M}$, respectively.

$$S_\alpha = \alpha \times S_o + (1-\alpha) \times S_r \quad (17)$$

It is utilized for the evaluation of spatial coherence and structural congruence between the region of interest and the GT. Where $S_o$ represents the similarity of the target structure, $S_r$ represents the similarity of the regional structure, α is the assigned weight of $S_o$ and $S_r$ which is set to

0.5, referring to [77].

$$F_\beta = \frac{(1+\beta^2) \times Precision \times Recall}{\beta^2 \times Precision + Rec} \quad (18)$$

It is a weighted harmonic of precision and recall, where the value of $\beta^2$ is equal to 0.3 as suggested in [78].

### C. Comparison with State-of-the-Art Methods

In this section, the proposed method is compared with 17 state-of-the-art methods on three datasets, including two traditional methods (i.e., DSG [13], RCNN [17]), seven NSI-SOD methods (i.e., SCRN [69], CPDNet [55], GateNet [70], CTDNet [71], PFSNet [72], BSANet [73], PGNet [74]), and eight ORSI-SOD methods (i.e., FSMINet [39], ERPNet [43], CorrNet [38], EMFINet [27], JRBM [42], SRAL [75], SeaNet [26]). Considering the fairness of the comparison, saliency maps of different models can be obtained by executing the source codes or provided by the authors. Since the authors of the LVNet [9] and SRAL [75] methods do not provide source code, we directly use the metrics from the original paper. Meanwhile, we retain NSI-SOD methods to generate the results by implementing with the recommended parameter settings.

*1) Quantitative Analysis:* To comprehensively and rigorously evaluate the performance of our proposed method, we conducted a quantitative analysis using three approaches: a) basic analysis of curves and metrics, b) attribute-based analysis on the ORSI-4199 dataset, and c) stability analysis.

*a) Curves and metrics analysis:* We generated F-measure curves and Precision-Recall (P-R) curves to evaluate the performance of the saliency model. In the F-measure curve, a larger area enclosed by the curve and the axis indicates better model performance. In the P-R curve, the curve being closer to the (1, 1) coordinate indicates better performance. As depicted in Fig. 9, the performance of our method on the three datasets is the best in both the F-measure curve and the P-R curve.

Moreover, to further intuitive representation, as reported in Table I, our method shows strong competitiveness and superiority on the tested three datasets. On the ORSSD dataset, our method exhibits slightly lower performance than SRAL and CorrNet in terms of the $F_\beta$ score. However, our method outperforms both SRAL and CorrNet in other evaluation metrics. ($\mathcal{M}$: 0.0093 (Ours) vs. 0.0105 (SRAL) and 0.0098 (CorrNet); $S_\alpha$: 0.9411 (Ours) vs. 0.9305 (SRAL) and 0.9380 (CorrNet)). On the EORSSD dataset, although our method does not perform as well as SRAL in two metrics, our method demonstrates competitive advantages in terms of the $S_\alpha$ score and ranks second among all methods in other metrics. In addition, SRAL utilizes super-resolution to aid learning and adopts a multitasking framework, whereas our approach does not have any additional task aids. On the most challenging ORSI-4199 dataset, our method surpasses all other methods in all evaluation metrics and pulls off a large gap. Compared with the second and third places, our method has improved the $S_\alpha$ score by 0.0167, 0.0172, the $F_\beta$ score by 0.0145, 0.0181, and the $\mathcal{M}$ score reduces by 0.0039, 0.0052. Additionally, the table also records the FLOPs, parameters (params.), model size, and frames per second (FPS) for various publicly available methods. To ensure a fair comparison, the same tensor input of size (1, 3, 256, 256) is used to obtain these metrics. Upon comparison, it is observed that while our method may not have the best performance in the four metrics, it consistently maintains a relatively neutral position, striking a better balance between these metrics.

*b) Attribute-based analysis:* The ORSI-4199 dataset provides nine scene patterns for a more intuitive comparison. We present the SSIM scores of BSCGNet and 6 SOTA NSI-SOD methods in Table II. Upon comparing Table I and Table II, it can be found that some methods, such as EMFINet, CorrNet's performance in the entire ORSI-4199 dataset is not the most prominent, but the performance in some scenarios can reach the top three.

TABLE II
ATTRIBUTE-BASED PERFORMANCE ON THE ORSI-4199 DATASET. THE AVERAGE SSIM SCORES FOR PARTICULAR ATTRIBUTES ARE PRESENTED. THE AVG. ROW REPORTS THE AVERAGE RESULTS FOR NINE ATTRIBUTES, AND TOP THREE SCORES IN EACH LINE ARE MARKED IN **RED**, **BLUE**, AND **GREEN**, RESPECTIVELY

| Attr. | FSMINet | ERPNet | CorrNet | EMFINet | JRBM | SeaNet | Ours |
|---|---|---|---|---|---|---|---|
| BSO | 0.7540 | **0.9051** | 0.8547 | **0.9195** | 0.8802 | 0.8809 | **0.9185** |
| CS | 0.7558 | **0.8904** | 0.8723 | **0.8990** | 0.8799 | 0.8836 | **0.9115** |
| CSO | 0.7370 | **0.8894** | 0.8320 | **0.9087** | 0.8569 | 0.8612 | **0.9148** |
| ISO | 0.7202 | **0.8884** | 0.8203 | **0.9075** | 0.8623 | 0.8548 | **0.9037** |
| LSO | 0.7486 | 0.8261 | **0.8383** | 0.8214 | 0.8336 | **0.8460** | **0.8470** |
| MSO | 0.7580 | 0.8533 | **0.8675** | 0.8494 | 0.8317 | **0.8591** | **0.8816** |
| NSO | 0.7189 | **0.8643** | **0.8778** | 0.8408 | 0.8478 | 0.8837 | **0.8837** |
| OC | 0.7470 | 0.8174 | **0.8654** | 0.8372 | 0.8376 | **0.8503** | **0.8762** |
| SSO | 0.7338 | 0.8196 | **0.8485** | 0.8192 | 0.8091 | **0.8331** | **0.8602** |
| Avg. | 0.7415 | **0.8616** | 0.8530 | **0.8670** | 0.8488 | 0.8614 | **0.8886** |

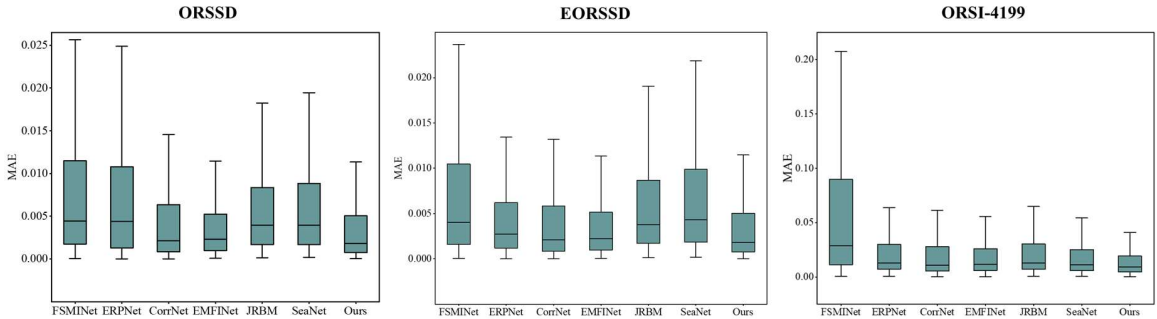

**Fig. 10.** Box plots of the MAE distributions for different RSI-SOD methods on the three datasets.

Furthermore, SeaNet achieves favorable results in Table I, but it does not perform as well as EMFINet in different scenarios. The above fact reflects that the performance advantages and disadvantages of different models for different remote sensing scene attributes are not reflected by the metrics on the whole test set [75]. It is worth noting that our method has a clear advantage in all 7 scenarios, while it is only slightly behind the second place in the remaining 2 scenarios, and the performance gap between the two is minimal. Most intuitively, the average score of our method in all scenarios is the highest, surpassing the second-place method by 0.0216, which effectively proves the superiority of our method over other ORSI-SOD methods in challenging scenarios, highlighting its strong stability.

*c) Stability Analysis:* To verify the stability of our method, we employed box plots, which are statistical graphs capable of visually representing the dispersion of a dataset. Box plots can accurately and stably represent the distribution of the data without being affected by the dispersion value. It has a better effect in evaluating the stability of the method, and the smaller the space of the box type, the more concentrated the data. Fig. 10 records the distribution of $\mathcal{M}$ score of our method and other 6 SOTA NSI-SOD methods on three datasets. It can be seen from the figure that the stability of our method on the three datasets has certain competitiveness and advantages compared with other NSI-SOD methods, especially in the ORSI-4199 dataset, the stability of our method is more obvious, which further confirms the above analysis of Table II.

*2) Qualitative Analysis:* To provide a more intuitive demonstration of our method's performance for ORSI-SOD, we conducted a detailed qualitative analysis on two datasets.

On the EORSSD dataset, we show five representative and challenging ORSI scenarios, which are presented in Fig. 11. They are specifically summarized as follows: (1) Three cases of buildings: general building, building with inconsistent colors, and multiple buildings; (2) Three cases of cars: multiple cars, cars with shadows, and cars with complex background; (3) Two cases of planes: general airplanes and multiple airplanes; 4) Two cases of rivers: river with low contrast and river with irregular topology; (5) Two cases of pools: pool with interferences and pool with complex background.

On the ORSI-4199 dataset, we further categorize the nine challenging attributes into six, which are presented in Fig. 12. The six challenging attributes are: (1) BSO: big salient object; (2) ISO: incomplete salient object; (3) OC: off center salient object; (4) CS and CSO: complex scene and complex salient object; (5) LSO and NSO: low contrast and narrow salient object; (6) SSO and MSO: small and multiple salient objects.

*a) Visualization comparison on the EORSSD dataset:* In terms of building detection, for buildings with three different attributes, most of the methods can locate the buildings very well, but many methods still have false detections and missed detections, mistaking cars and roads as salient objects. (CPDNet, PGNet, and EMFINet, etc.), whereas our method not only accurately locates the salient objects, but also has finer details of the silhouettes. In the detection of cars and airplanes, regardless of the specific attributes of the targets, one common characteristic is small-sized objects. However, the interference of small objects and other non-target objects or environments can lead to the dilution of semantic information and a decrease in semantic relevance. Nevertheless, our method produces salient maps that exhibit a remarkable resemblance to the GT, demonstrating a higher level of completeness. In the detection of rivers and swimming pools, our method exhibits richer details compared with other methods. It also demonstrates enhanced refinement in handling boundaries. Specifically, in the case of river detection, the low contrast and irregular interferences often blur the contour details and semantic expressions of the target, leading to suboptimal detection results in most methods. However, our method achieves a higher degree of completeness compared to other methods. The excellent performance of our method will be more prominent on the more challenging ORSI-4199 dataset.

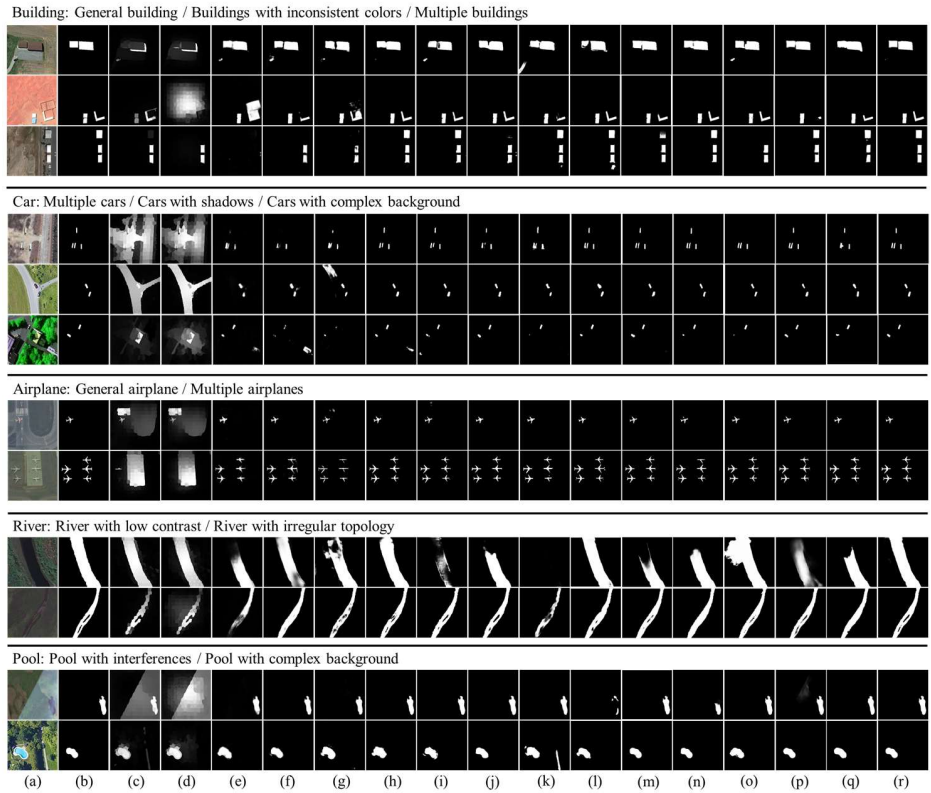

**Fig. 11.** Qualitative analysis of different saliency methods on EORSSD dataset. (a) Optical RSIs. (b) GT. (c) DSG. (d) RCNN. (e) SCRN. (f) CPDNet. (g) GateNet. (h) CTDNet. (i) PFSNet. (j) BSANet. (k) PGNet. (l) FSMINet. (m) ERPNet. (n) CorrNet. (o) EMIFNet. (p) JRBM. (q) SeaNet. (r) Ours. *Please zoom-in for the best view.*

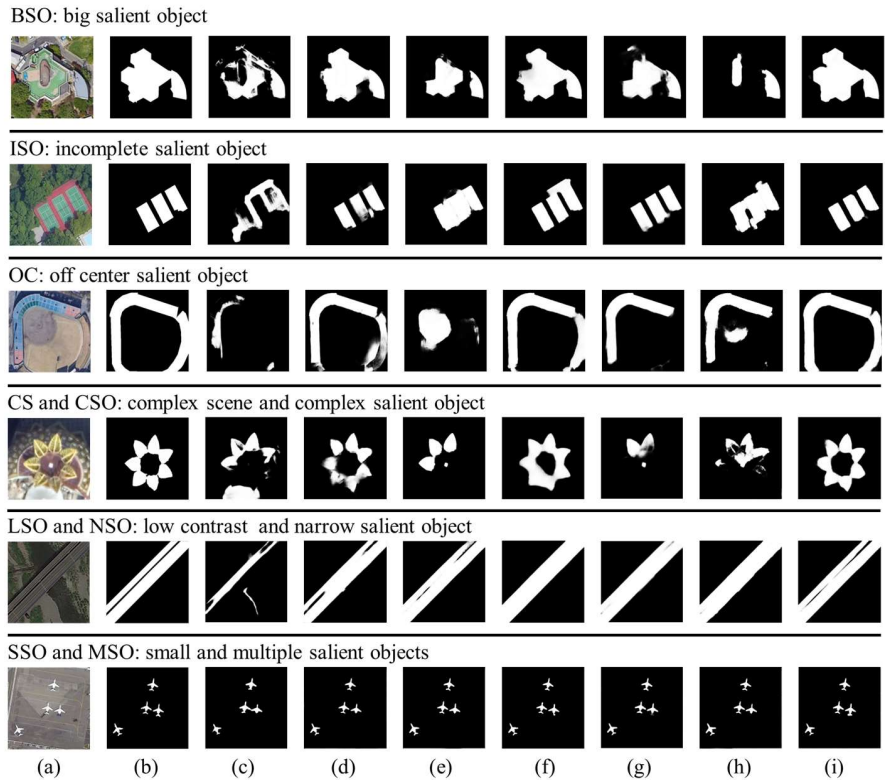

**Fig. 12.** Qualitative analysis of RSI-SOD models on more challenging scenarios of ORSI-4199 dataset. (a) Optical RSIs. (b) GT. (c) FSMINet. (d) ERPNet. (e) CorrNet. (f) EMIFNet. (g) JRBM. (h) SeaNet. (i) Ours. *Please zoom-in for the best view.*

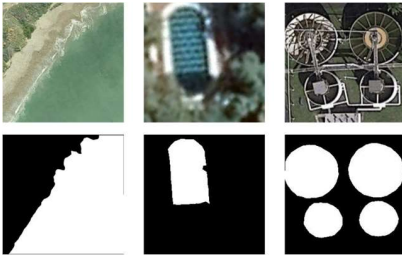

**Fig. 13.** Some representative samples: (top row) optical RSIs and (bottom row) GT.

*b) Visualization comparison on the ORSI-4199 dataset:* We can further summarize the results in Fig. 12 into four types of interference for detailed analysis, namely (1) weak semantic correlation (BSO, OC, SSO and MSO); (2) strong semantic correlation (LSO and NSO); (3) The target is occluded (ISO); (4) Complex background environment (CS and CSO).

First, in the images presented by BSO, OC, SSO and MSO, a common observation can be made: the same target exhibits non-uniform or significant variations in color appearance. This characteristic tends to weaken the semantic correlation between classes, which can misguide the model's learning process and result in missed detections. For example, in the cases of BSO and OC, it can be observed that most ORSI-SOD methods, except for ERPNet, EMFINet and our method, exhibit significant omissions or misdetections on a large scale.

Similarly, in SSO and MSO, nearly all methods fail to capture the tails of small airplane, but our proposed method addresses the weak semantic relevance by achieving superior completeness in the results. In contrast, in the images demonstrated by LSO and NSO, the presence of low contrast interference amplifies the semantic correlation between the gap in the bridges and the neighbouring bridges. If the model fails to accurately recognize the distinctive features of the targets, it may incorrectly classify them as the same class of objects. For example, with the exception of our method, CorrNet, and ERPNet, none of the other methods are able to distinguish gaps from bridges. Second, in the case of ISO, the occlusion of salient targets introduces a bias in the model's boundary localization. From the figure, it is evident that only our method and EMFINet accurately localize the boundaries of the occluded regions. However, our method exhibits superior boundary details on the other boundaries of the targets compared with EMFINet. Lastly, in CS and CSO, the presence of a complex background with similar colors to the salient target poses challenges in accurately localizing the salient object. However, our method outperforms other methods in detecting salient targets with better and richer details in these scenarios.

### D. The Validation of Model Transferability

To verify the generalization ability of the model, we conducted a transferability analysis. In this section, we use a new dataset (i.e., ORISOD dataset) constructed by Zheng et al. [45] to verify the transferability of the ORSI-SOD method. This is the first time that this dataset has been used for the generalization study of the ORSI-SOD method. To better judge the transferability of each model, we refrained from training them from scratch. Instead, we directly applied the trained weight obtained from the ORSI-4199 dataset to the ORISOD dataset. This dataset has 3784 images with corresponding labels in the train set and 1270 images with corresponding labels in the test set. To make the experiment more challenging, we merged the train and test sets so that there are a total of 5054 images for us to test. Notably, the style and type of salient objects depicted in these images differ somewhat from those in the dataset mentioned in section IV(A) 1), some representative samples are shown in Fig. 13. Therefore, such verification process is very challenging.

Table III presents the performance of all ORSI-SOD methods on the ORISOD dataset. Upon examination of the table, it is evident that our method scores the highest on all three metrics. Among them, the $\mathcal{M}$ score is reduced by 0.11%-1.47%, the $S_\alpha$ score and $F_\beta$ score are improved by 0.34%-4.33%, 0.69%-7.08%. Overall, our method exhibits superior generalization ability compared with other ORSI-SOD methods, further validating its superiority in the field.

### E. Ablation Analysis

To further demonstrate the effectiveness of key components, this section presents detailed ablation experiments. Specifically, we conduct five groups of experiments, investigating the composition of different components and the impact of different loss functions. Furthermore, we delve deeply into the genuine contributions brought by the proposed modules. Table IV records the metrics on the ORSSD and EORSSD datasets under different conditions. It is worth noting that each experiment in this section employs the same parameter settings as in section IV(A) 2) and undergoes retraining.

*1) W/ BPC module:* When the BPC module (No.2) is added on top of the baseline (No.1), on the ORSSD dataset, $\mathcal{M}$ decreased by 0.43%, while $S_\alpha$ and $F_\beta$ increased by 1.9% and 1.89%, respectively. On the EORSSD dataset, $\mathcal{M}$ decreased by 0.06%, while $S_\alpha$ and $F_\beta$ increased by 1.19% and 2%, respectively. This demonstrates that the calibration and preservation of boundary cues contribute to accuracy improvement. Moreover, the BPC module introduces only 0.32M parameters.

TABLE III
TRANSFERABILITY VERIFICATION ON THE ORISOD DATASET. HERE, "↑"(↓) MEANS THAT THE LARGER (SMALLER) THE BETTER. THE BEST THREE RESULTS IN EACH ROW ARE MARKED IN **RED**, **BLUE**, AND **GREEN**, RESPECTIVELY

| Metrics | FSMINet | ERPNet | CorrNet | EMIFNet | JRBM | SeaNet | Ours |
|---|---|---|---|---|---|---|---|
| $\mathcal{M}\downarrow$ | 0.1096 | 0.1138 | **0.1053** | 0.1168 | 0.1068 | **0.1002** | **0.0991** |
| $S_\alpha\uparrow$ | 0.7020 | 0.7196 | **0.7325** | 0.7282 | 0.7297 | **0.7419** | **0.7453** |
| $F_\beta\uparrow$ | 0.6101 | 0.6394 | **0.6640** | 0.6534 | 0.6440 | **0.6740** | **0.6809** |

TABLE IV
ABLATION EXPERIMENTAL RESULTS OF DIFFERENT COMPONENTS AND LOSS FUNCTIONS. "B" AND "I" REPRESENT BCELOSS AND IOULOSS RESPECTIVELY. "↑"(↓) MEANS THAT THE LARGER (SMALLER) THE BETTER. THE BEST RESULT IS **BOLD**

| No. | Baseline | BPC | DFFC | AFR | Loss | ORSSD $\mathcal{M}\downarrow$ | ORSSD $S_\alpha\uparrow$ | ORSSD $F_\beta\uparrow$ | EORSSD $\mathcal{M}\downarrow$ | EORSSD $S_\alpha\uparrow$ | EORSSD $F_\beta\uparrow$ | Params(M)↓ | FLOPs(G)↓ |
|---|---|---|---|---|---|---|---|---|---|---|---|---|---|
| 1 | ✓ | | | | B+I | .0201 | .8952 | .8557 | .0115 | .9029 | .8392 | 25.69 | 59.53 |
| 2 | ✓ | ✓ | | | B+I | .0158 | .9142 | .8746 | .0109 | .9148 | .8592 | 26.01 | 78.78 |
| 3 | ✓ | | ✓ | | B+I | .0133 | .9202 | .8820 | .0101 | .9150 | .8597 | 25.79 | 60.05 |
| 4 | ✓ | | | ✓ | B+I | .0166 | .9133 | .8758 | .0110 | .9090 | .8522 | 26.63 | 70.78 |
| 5 | ✓ | ✓ | ✓ | ✓ | B | .0109 | .9295 | .9038 | .0105 | .9153 | .8699 | 26.99 | 86.35 |
| 6 | ✓ | ✓ | ✓ | ✓ | B+I | **.0093** | **.9411** | **.9101** | **.0079** | **.9296** | **.8808** | 26.99 | 86.35 |

From Fig. 14, we can observe the contributions brought about by the BPC module. Based on the preliminary analysis of the BPC module in section III(B), it can be inferred that this module effectively preserves the position information of edges in low-level features. By employing a level-by-level optimization approach, it successfully mitigates the issue of edge information loss during forward propagation. As depicted in Fig. 14, (a1) - (a3) represent the second to fourth layers of the encoder, i.e., $E_i \{i=2, i=3, i=4\}$ in section III(B), while (b1) - (b3) denote the feature maps processed by the BPC module. It is evident from (a1) - (a3) that during the forward propagation of the encoder, the boundary position information becomes increasingly blurred and generates a significant amount of noise, and particularly in (a3), the edge information has almost been entirely lost. And after going through the BPC module, the boundary information is protected to some extent in the process of layer-to-layer transmission due to the effective utilization of adjacent features. In addition, the FC sub-module also reduces the noise around salient objects. These optimizations are clearly reflected in (b1) - (b3). Specifically, in (b1) and (b2), the edge information is more accurate and the ambient noise is reduced compared with before calibration (i.e., (a1) and (a2)). In (b3), the contribution of the BPC module is more prominent, and the boundary information is clearly protected. Overall, these visualized feature maps strongly demonstrate the true contribution of the BPC module.

*2) W/ DFFC module:* With the inclusion of the DFFC module (No.3), all three metrics showed improvement on both datasets.

On the ORSSD dataset, $\mathcal{M}$ decreased by 0.68%, while $S_\alpha$ and $F_\beta$ increased by 2.5% and 2.63%, respectively. On the EORSSD dataset, $\mathcal{M}$ decreased by 0.14%, while $S_\alpha$ and $F_\beta$ increased by 1.21% and 2.05%, respectively. It is noteworthy that, even with the introduction of very low parameters and complexity, such superior performance is maintained, effectively demonstrating the importance and effectiveness of this module.

Fig. 15 illustrates the contributions brought about by the DFFC module. The DFFC module, which supplements the features of each layer of the encoder (i.e., $E_1 - E_5$) through boundary-semantic dual features, is analyzed in section III(C). As shown in Fig. 15, the obtained dual-feature map (i.e., the Feedback Map in the figure) not only contains rich semantic information but also prevents the boundary information of salient objects from being entirely lost. The attention map with dual features is fed back to each layer of the encoder to supplement the required features and obtain more richer feature maps, as shown in $G_1 - G_5$. Specifically, for the lower-level features $G_1 - G_3$, the DFFC module not only preserves the edge position information but also suppresses noise while improving the brightness of the internal region of salient objects. In other words, the DFFC module enhances the semantic information inside the salient objects. In $G_4$, the DFFC module also suppresses some non-target noise but weakens the semantic information of the salient objects. This is believed to be caused by the ambiguous semantic and localization information about salient objects in $E_4$, we believe that this is because the semantic and location of salient objects are relatively vague, which makes the SELayer uncertain about the target's attention and only preserve weak position information. The last layer of the encoder, $E_5$ possesses the strongest semantic information.

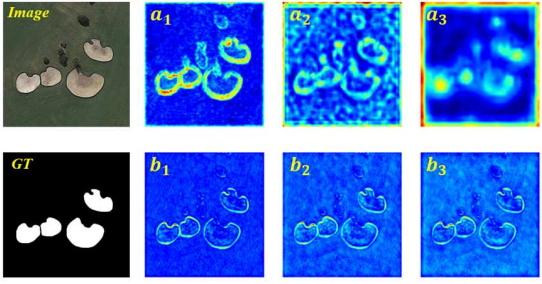

**Fig. 14.** Illustration of the real contribution of the BPC module.

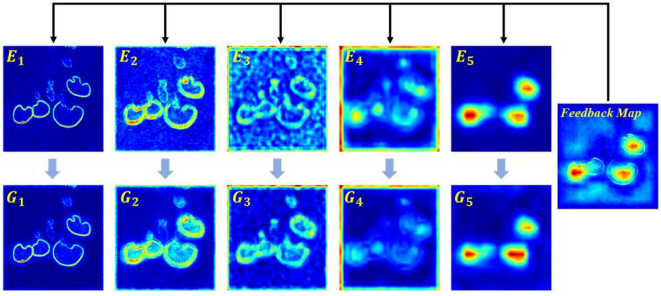

**Fig. 15.** Illustration of the real contribution of the DFFC module.

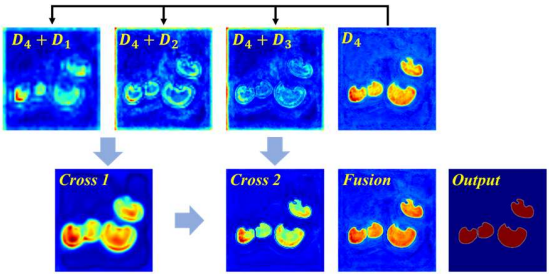

**Fig. 16.** Illustration of the real contribution of the AFR module.

When the feedback is received, the boundary position information in the feedback map makes the shape of the salient objects closet to the GT, as shown in $G_5$. Therefore, the real contribution of the DFFC module is effectively demonstrated through visual feature map presentation

*3) W/ AFR module:* Similarly, when introducing the AFR module (No.4), all metrics showed improvements. On the ORSSD dataset, $\mathcal{M}$ decreased by 0.35%, while $S_\alpha$ and $F_\beta$ increased by 1.81% and 2.01%, respectively. On the EORSSD dataset, $\mathcal{M}$ decreased by 0.05%, while $S_\alpha$ and $F_\beta$ increased by 0.61% and 1.3%, respectively. From the data, it can be observed that this module brings relatively fewer improvements compared with the other modules. We attribute this to the fact that this module operates on the final layer of the network. Specifically, because the last layer is typically close to the real labels, most existing methods use it directly as the final output. This approach also yields good accuracy. However, we contemplate further processing on this foundation, using it to refine and enrich certain more detailed regions. As a result, the extent of accuracy improvement it offers is relatively limited.

Similarly, from Fig. 16, we can clearly observe the contributions brought about by the AFR module. In section

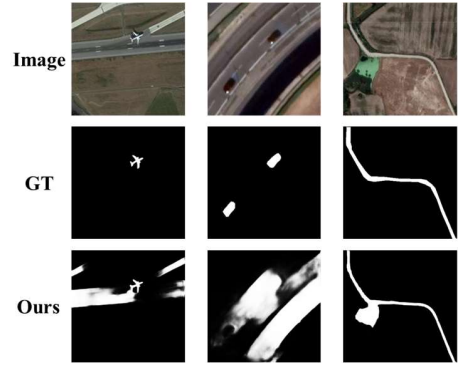

**Fig. 17.** Some representative failure cases.

III(D), the AFR module is introduced to obtain more accurate saliency maps. Fig. 16 shows that as the direct output of the last layer of the decoder, $D_4$ is highly similar to the GT. So, we fed it back to each layer of the decoder and used the submodules described in section III(D) 2) for feature refinement. "Cross 1" in the figure is the result of crossing "$D_4 + D_1$" and "$D_4 + D_2$" after cross refinement. It can be seen that the salient object is more accurately located at this stage, but the accurate judgment of the boundary is still lacking. "Cross 2" is the result of further crossing "$D_4 + D_3$" and "Cross 1". Although we have a clearer judgment of the boundary this time, we have a lower degree of confidence in the interior of salient objects. These problems have been solved through information fusion output, and as seen in the figure, the result of "Fusion" is more accurate than $D_4$. The final saliency map is shown as "Output" in the figure. Above all, the analysis of the visualized feature maps strongly proves the necessity of the AFR module.

*4) W/o IoULoss:* Regarding the different combinations of loss functions, we exclusively employed the standalone BCELoss in No. 5. Due to the unavailability of IoULoss for global evaluation, there is a certain decrease in the metrics.

Overall, in this section, we have conducted a comprehensive quantitative and visual analysis of the proposed modules, further substantiating the effectiveness and rationality of the method presented in this paper, thereby offering substantial performance improvements to ORSI-SOD.

### F. Failure Cases and Limitations

Section IV(C) to IV(E) fully prove the rationality and effectiveness of the method BSCGNet proposed in this paper for ORSI-SOD. However, it is important to acknowledge that our method still has limitations when applied to specific challenging scenarios. Fig. 17 presents several representative examples of these challenging scenarios.

From the three sets of images, it can be found that each image contains not only one type of salient object in the dataset category, for example: (a) contains airplanes and roads; (b) contains cars, roads and low-contrast rivers; (c) contains roads and lakes. It is worth noting that the training set of the original data already includes these objects, and the amount of data is not small yet, in other words, it is likely that these objects may also appear in other scenarios as salient object properties. Therefore, during model training, the model has learned the semantic features of these objects, which can result in the presence of two or more salient objects with different attributes. However, it is rare to encounter images in the dataset where two salient objects with different attributes are labeled together, which will cause a deviation between the results obtained by our method and the GT.

On the other hand, as can be seen from Table I, the metrics of our method on small datasets (i.e., ORSSD and EORSSD datasets) are not all the best, but it demonstrates an absolute advantage on large and challenging datasets like ORSI-4199 dataset. We attribute this to the fact that our method primarily focuses on optimizing and enhancing details and semantic information. However, due to the limited size of the dataset, it may not provide sufficient samples and diversity to comprehensively cover various scenarios and variations, making it difficult to capture enough semantic information and details.

For the particular challenging scenario described above, in future work we will consider domain adaptation to help the model better adapt to deviations between data. Additionally, to overcome the performance limitations on small datasets, we will investigate novel data enhancement strategies to increase the diversity of datasets. Alternatively, we will explore the development of lightweight model architectures that are specifically designed to perform well on small datasets.

## V. CONCLUSION

In this paper, we investigate the importance of boundary cues for salient object detection and explore the significance of the last layer of the decoder. Subsequently, we propose a novel dedicated method called BSCGNet for ORSI-SOD. Considering the loss of boundary information in high-level features, we introduce the BPC module to safeguard and calibrate the boundary cues, simultaneously suppressing noise in low-level features. In the encoder and decoder, we also adopt two unique feedback mechanisms and propose the DFFC module and AFR module. The former aims to coordinate the features of each layer in the encoder and achieve feature supplementation, while the latter is used in the decoder to eliminate the feature differences between layers and obtain more refined saliency maps. Finally, extensive experiments, including design rationalization verification of key components, qualitative and quantitative analysis, the validation of model transferability, computational efficiency analysis, ablation study, and failure cases analysis, strongly demonstrate that the BSCGNet outperforms 17 SOTA methods and significantly improves the performance of ORSI-SOD.